\documentclass[10pt,twocolumn,letterpaper]{article}

\usepackage[pagenumbers]{cvpr}

\usepackage{amsmath}
\usepackage{amssymb}
\usepackage{bbding}
\usepackage{booktabs}
\usepackage{floatrow}
\usepackage{graphicx}
\usepackage{multirow}
\usepackage{nccfoots}
\usepackage{tabularx}
\usepackage{threeparttable}
\usepackage[colorlinks,pagebackref]{hyperref}

\def\confName{CVPR}
\def\confYear{2022}
\def\cvprPaperID{4076}

\newcommand{\email}[2]{\href{mailto:#1}{\textcolor{black}{#2}}}
\newcommand{\authorfootnote}[1]{\Footnotetext{}{\textit{$^*$#1}}}

\makeatletter
\def\@hangfrom#1{\setbox\@tempboxa\hbox{{#1}}
\hangindent 7mm
\noindent\box\@tempboxa}
\makeatother

\begin{document}

\title{UMT: Unified Multi-modal Transformers for Joint Video Moment \\ Retrieval and Highlight Detection}

\author{Ye~Liu\,$^1$\quad Siyuan~Li\,$^2$\quad Yang~Wu\,$^{2*}$\quad Chang~Wen~Chen\,$^{1,4}$\quad Ying~Shan\,$^2$\quad Xiaohu~Qie\,$^3$\\
$^1$\,Department of Computing, The Hong Kong Polytechnic University \\
$^2$\,ARC Lab, Tencent PCG\quad $^3$\,Tencent PCG\quad $^4$\,Peng Cheng Laboratory \\
{\tt\small\email{csyeliu@comp.polyu.edu.hk}{csyeliu@comp.polyu.edu.hk}, \email{changwen.chen@polyu.edu.hk}{changwen.chen@polyu.edu.hk}} \\
{\tt\small\{\email{siyuanli@tencent.com}{siyuanli},\email{dylanywu@tencent.com}{dylanywu},\email{yingsshan@tencent.com}{yingsshan},\email{tigerqie@tencent.com}{tigerqie}\}@tencent.com}}

\twocolumn[{
\maketitle
\centering
\vspace{-0.4cm}
\includegraphics[width=0.9\linewidth]{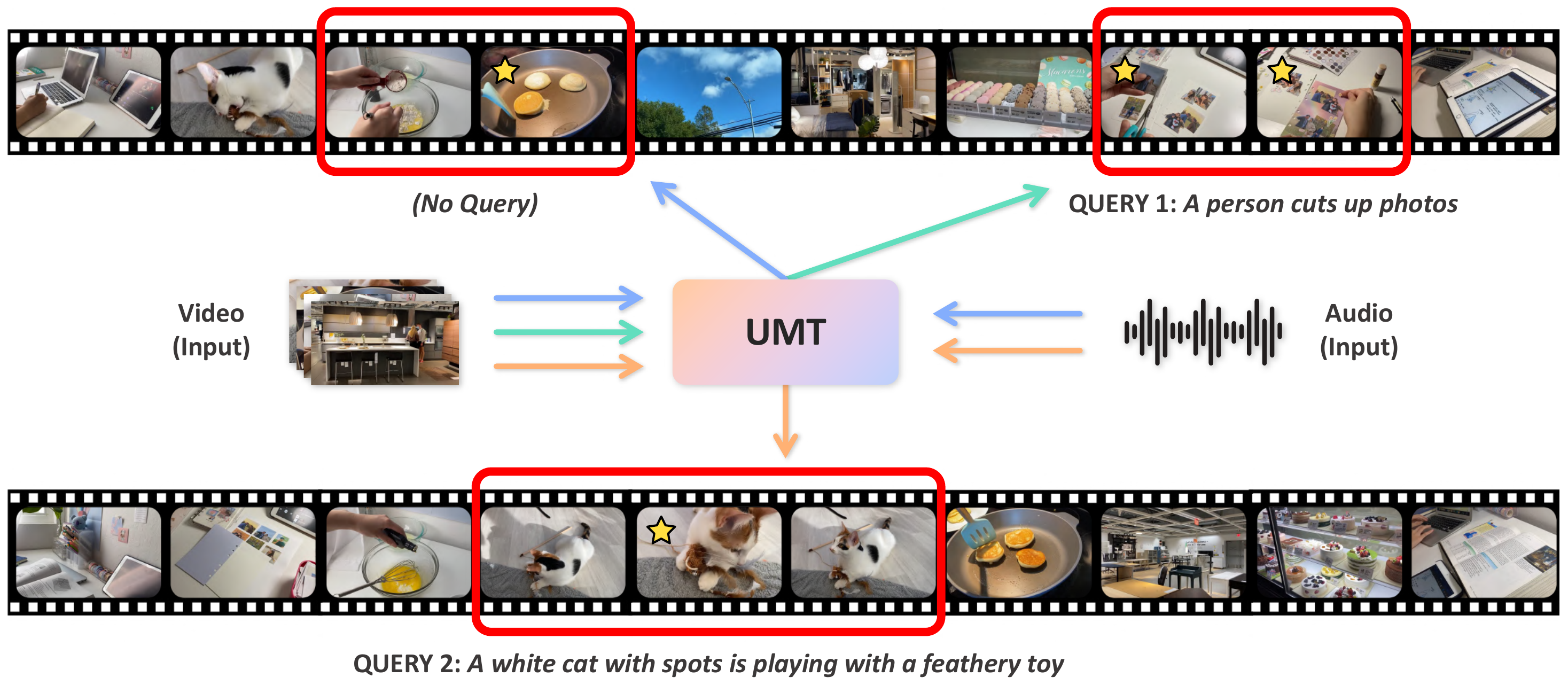}
\captionof{figure}{The proposed UMT is a unified and flexible framework which can handle different input modality combinations, and output video moment retrieval and/or highlight detection results (marked by red rectangles and golden stars, respectively). Note that different text queries lead to different outcomes from the same video. Arrows in different colors denote different input-output combinations.}
\vspace{0.6cm}
\label{fig:1}}]

\authorfootnote{Corresponding author.}

\begin{abstract}

Finding relevant moments and highlights in videos according to natural language queries is a natural and highly valuable common need in the current video content explosion era. Nevertheless, jointly conducting moment retrieval and highlight detection is an emerging research topic, even though its component problems and some related tasks have already been studied for a while. In this paper, we present the first unified framework, named Unified Multi-modal Transformers (UMT), capable of realizing such joint optimization while can also be easily degenerated for solving individual problems. As far as we are aware, this is the first scheme to integrate multi-modal (visual-audio) learning for either joint optimization or the individual moment retrieval task, and tackles moment retrieval as a keypoint detection problem using a novel query generator and query decoder. Extensive comparisons with existing methods and ablation studies on QVHighlights, Charades-STA, YouTube Highlights, and TVSum datasets demonstrate the effectiveness, superiority, and flexibility of the proposed method under various settings. Source code and pre-trained models are available at \href{https://github.com/TencentARC/UMT}{https://github.com/TencentARC/UMT}.

\end{abstract}

\vspace{-0.3cm}

\section{Introduction}\label{sec:1}

Video has already become the major media in content production, distribution, and consumption in our daily lives. It has the unique advantage of being able to include visual, audio, and linguistic information in the same media, in line with our natural experiences. Such an advantage on information richness, however, is also a challenging factor limiting its production and consumption, as it brings about very high costs on satisfying two critical needs. The first one is to find relevant moments in existing videos for producing new content or just getting creation hints from such references. The second one is to glance at large amounts of video content quickly by scanning video highlights rather than watching the entire original videos or video moments at a normal speed, which is needed by both video producers and consumers in such a content explosion era.

These two critical needs lead to two important research topics: video moment retrieval \cite{anne2017localizing,gao2017tall} and video highlight detection \cite{wang2004sports,sun2014ranking,yao2016highlight}, respectively. Although one may realize that these two tasks are closely related (especially when a text query is given), they have not yet been explicitly jointly studied until a very recent work \cite{lei2021qvhighlights} which builds a novel dataset called QVHighlights for this purpose and presents the first model Moment-DETR optimized for jointly solving both problems. Nevertheless, this seminal work has several limitations. It assumes a text query always exists and it has only applied to the visual modality of each video. Moreover, it is still a very basic model called a strong baseline, although it adopts a transformer framework, the latest and fast-rising neural network architecture type.

This paper goes deeper into designing joint video moment retrieval and highlight detection approaches by mainly exploring two aspects: \emph{multi-modal learning} and \emph{flexibility}, as shown in Figure~\ref{fig:1}. Apart from text and video (\ie visual information), audio is also treated as an important input. Moreover, a unified yet flexible framework called Unified Multi-modal Transformers (UMT) is proposed to handle different modality reliability situations and combinations. For example, when the text input is unavailable, the task degenerates to highlight detection only. When there is some significant distraction in the text, its reliability will be compromised. Moreover, the audio may also be noisy, which may limit effective exploration. UMT covers all these natural variations which conventionally need to be resolved by different specifically designed models. 

To demonstrate the effectiveness and superiority of the proposed framework, we conduct experiments not only on the QVHighlights dataset \cite{lei2021qvhighlights}, the only one built for joint video moment retrieval and highlight detection, but also on popular public datasets for moment retrieval (Charades-STA \cite{gao2017tall}) and highlight detection (TVSum \cite{song2015tvsum}, YouTube Highlights \cite{sun2014ranking}), with or without text guidance. For each case, we compare the proposed scheme with several state-of-the-art approaches. Detailed ablation studies are also carried out to evaluate the essential components of the proposed scheme and to reveal meaningful insights.

\begin{figure*}
\centering
\includegraphics[width=0.95\linewidth]{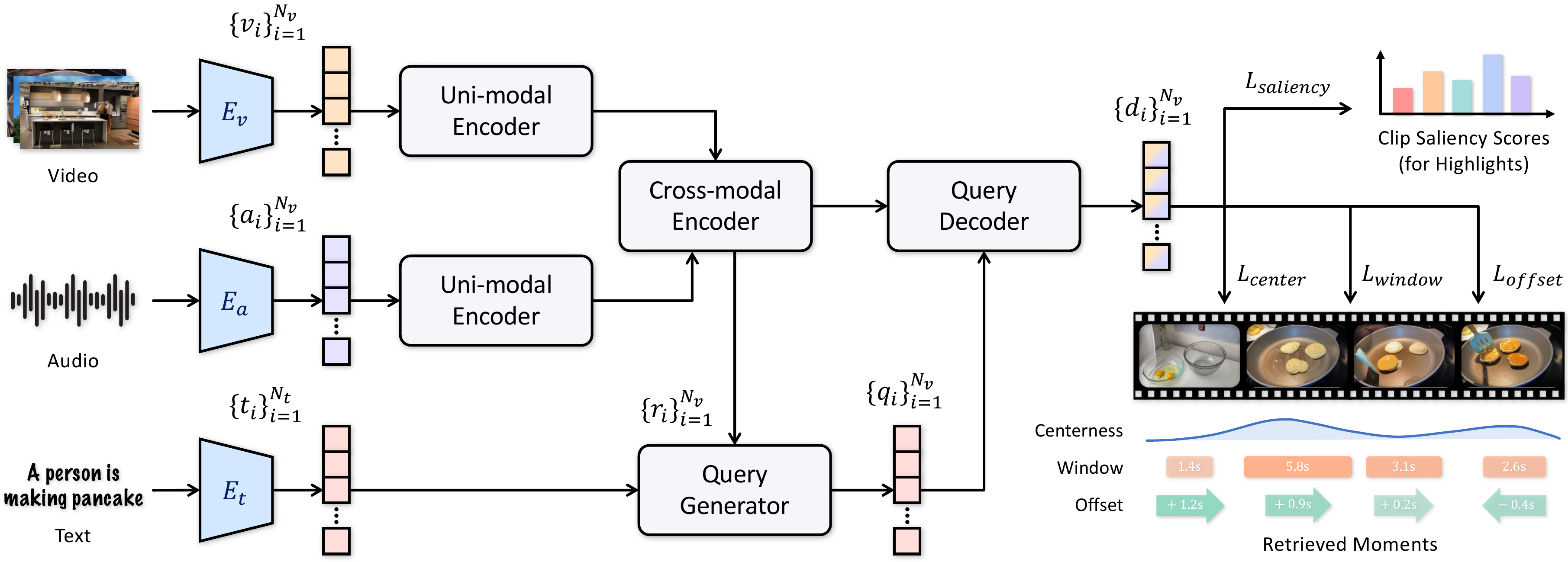}
\caption{Overall architecture of our framework. When either the video or audio is unavailable, the corresponding uni-modal encoder and cross-modal encoder are deactivated. If text queries are not provided, the model would simply use learnable moment queries instead. Detailed explanations of notations are described in Section~\ref{sec:3.1}.}
\label{fig:2}
\end{figure*}

\section{Related Works}

\paragraph{Video Moment Retrieval \& Highlight Detection}

Moment retrieval is a recently studied research topic that focuses on retrieving related moments in a video given a natural language query. Most existing works \cite{anne2017localizing,gao2017tall} assume there is only a single moment in a video corresponding to a given text query, and such queries are usually about activities. The recently proposed QVHighlights dataset \cite{lei2021qvhighlights} goes beyond that by annotating multiple moments in a video for each query and breaks the former moment distribution bias (locating more likely at the beginning of the videos). Video retrieval via text query \cite{otani2016learning} is a similar task, but it retrieves whole videos rather than video moments. Some works on language grounding \cite{tellex2009towards,regneri2013grounding} align textual phrases to temporal video segments, which operate at a finer scale than moment retrieval and target at different applications. Highlight detection concerns about detecting interesting or salient video segments (\ie highlights) in a video. It has a long history of about two decades with a rich literature, covering various domains of videos, including sports \cite{wang2004sports,hao2011detecting}, social media \cite{sun2014ranking}, and first-person \cite{yao2016highlight}. QVHighlights is the only dataset supporting highlight detection conditioned on text-guided moment retrieval results. Video summarization is a closely related task that targets at summarizing a long video with short video segments. It focuses on representativeness, diversity, and storyline, and thus it tends to be considered as a downstream application of highlight detection \cite{yao2016highlight}. Dynamic video thumbnail generation is another downstream task, which selects attractive video highlights and reforms them into a very short segment \cite{xu2021gif} to serve as the thumbnail. Among all these tasks, moment retrieval and highlight detection are two fundamental ones and they get highly correlated when a text query is given. This study follows the seminal work of QVHighlights on modeling both tasks together within a single framework. Different from Moment-DETR \cite{lei2021qvhighlights}, our model has the flexibility to perform moment retrieval or highlight detection only.

\paragraph{Text Query Based Models}

While text query is a must for moment retrieval, it seldom appears in the studies for video highlight detection, though we believe that providing text queries leads to better results as highlights are usually subjective and interest-dependent. An early work \cite{kudi2017words} proposes using text to find video highlights, but it is just about using a text ranking algorithm to rank video descriptions in the text domain for providing supervision to video shot ranking, not directly matching text and highlights. The only text-guided highlight detection exists in the very recent work \cite{lei2021qvhighlights}. In the closely related field of video thumbnail generation, text queries were first investigated in \cite{yuan2019sentence}, where a graph convolutional network is used to model the word-by-clip interactions. Later on, sentence-guided temporal modulation mechanism \cite{rochan2020sentence} is proposed to modulate an encoder-decoder based network. All these works assume the reliability of the text query and have to rely on it, while our proposed framework can easily work without text queries or with unreliable text queries.

\paragraph{Multi-modal Learning}

Recently, multi-modal learning approaches have been explored for highlight detection by jointly modeling visual and audio modalities. The earliest work seems to be MINI-Net \cite{hong2020mini}, which simply concatenates the feature vectors of both modalities. Very recently, two more sophisticated modality fusion models \cite{ye2021temporal,badamdorj2021joint} have been proposed and significantly boosted the performances. One of them invents a visual-audio tensor fusion mechanism \cite{ye2021temporal} to learn cross-modality relationships with tensor decomposition and low-rank constraints. The other does the fusion via cross-modal bidirectional attention layers \cite{badamdorj2021joint} which extract audio-attended visual features and visually-attended audio features. Though all the three approaches share the same idea with us on learning multi-modal fused representations for highlight detection, only the attention-based work build their model under the same supervised-learning setting as ours (the other two are weakly-supervised and thus not fairly comparable). Moreover, to the best of our knowledge, our approach is the first one to solve joint moment retrieval and highlight detection with multi-modal (visual-audio) learning.

\section{Method}

\subsection{Overview}\label{sec:3.1}

Given an untrimmed video $V$ containing $N_v$ clips and a natural language query $T$ with $N_t$ tokens, the goal of joint video moment retrieval and highlight detection is to localize all the moments (represented by temporal boundaries $b \in \mathbb{R}^2$) in $V$, in which the visual and/or audio contents are highly relevant to $T$, while predicting clip-level saliency scores $\{s_i\}_{i=1}^{N_v}$ for each moment simultaneously.

As shown in Figure~\ref{fig:2}, the overall architecture of our framework derives from the transformer encoder-decoder structure, and can be divided into five parts, \ie uni-modal encoder, cross-modal encoder, query generator, query decoder, and prediction heads. The input video and text are firstly processed by pre-trained feature extractors. Specifically, we use three different models ($E_v$, $E_a$, and $E_t$) to extract visual, audio, and textual features, respectively. Each video-text pair can be therefore represented by three collections of feature vectors, namely visual features $\{v_i\}_{i=1}^{N_v}$, audio features $\{a_i\}_{i=1}^{N_v}$, and textual features $\{t_i\}_{i=1}^{N_t}$. The visual and audio features are fed into separate uni-modal encoders to be contextualized under global receptive field, then be fused by the cross-modal encoder for visual-audio joint representations $\{r_i\}_{i=1}^{N_v}$. These representations, together with textual features, are used to generate clip-level moment queries $\{q_i\}_{i=1}^{N_v}$ that can be utilized to retrieve moments and highlights from joint representations in the query decoder. After decoding query-guided video features $\{d_i\}_{i=1}^{N_v}$, we use two prediction heads to obtain the final moment retrieval and highlight detection results.

\subsection{Uni-modal Encoder}

\begin{figure}
\centering
\includegraphics[width=\linewidth]{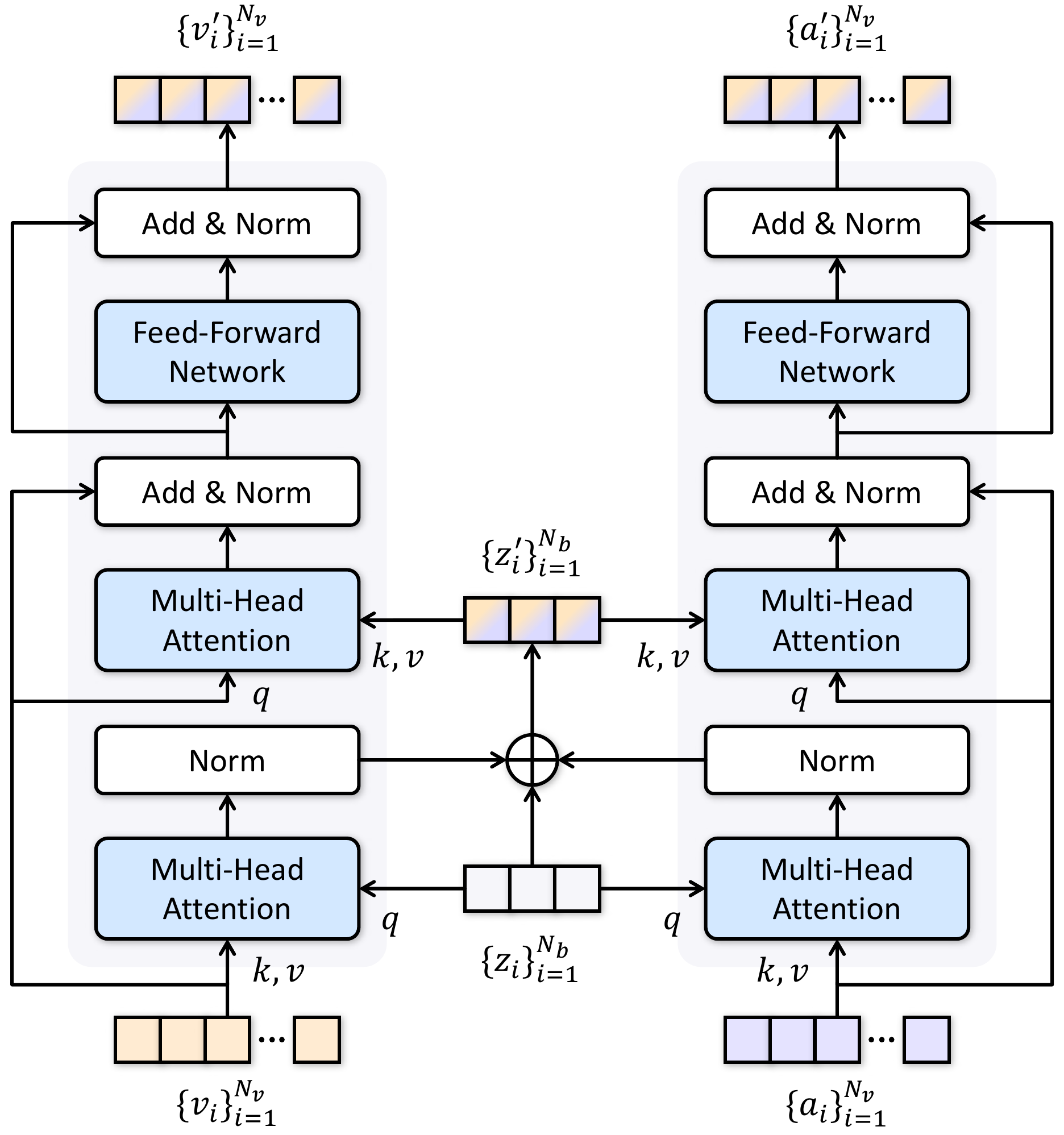}
\caption{The architecture of bottleneck transformer module. We introduce bottleneck tokens for cross-modal feature compression and expansion, largely reducing the computational cost.}
\label{fig:3}
\end{figure}

Most existing feature extractors for videos and audios \cite{tran2015learning,carreira2017quo,kong2020panns} are under the sliding window scheme, thus these methods only consider local temporal correlations, without being aware of the global context information, which is of great essence for video understanding tasks. Detecting queried moments and highlights in a video also requires an overall view of the global content. Therefore, to augment the features with global context within each modality, we adopt uni-modal encoder to process the input visual and audio features. This module is constructed by stacking standard transformer encoder layers \cite{vaswani2017attention}, each consisting of a multi-head self-attention block and a feed-forward network. In each attention head, self-attention for either of the visual or audio modality $x \in \{v, a\}$ can be computed as
\begin{gather}\label{eq:1}
x_i' = x_i + w_z \sum_{j=1}^{N_v} \frac{\exp(w_q x_i \times w_k x_j)}{\sum_{m=1}^{N_v} \exp(w_q x_i \times w_k x_m)} w_v x_j
\end{gather}
where $x_i$ and $x_i'$ are the input and output features of clip $i$, and $w_{\{q, k, v, z\}}$ indicates linear transform weights for the query, key, value, and output matrices. More details about self-attention computation are referred in \cite{vaswani2017attention}. The above formula computes the embedded gaussian correlations among clips, and aggregates the global context information into each clip. After aggregating the features, subsequently, a two-layer feed-forward network formed by \texttt{Linear\,$\rightarrow$\,ReLU\,$\rightarrow$\,Dropout\,$\rightarrow$\,Linear} is used to further project the features.

\subsection{Cross-modal Encoder}

Previous works \cite{badamdorj2021joint,nagrani2021attention} have claimed that jointly modeling multi-modal features can better obtain the overall representations. Hence after the uni-modal encoders, an extra cross-modal encoder is utilized to jointly capture the global correlations across modalities. Here, the exact form of the cross-modal encoder is not crucial. A straightforward approach is to apply cross-modal attention \cite{badamdorj2021joint}. However, such a strategy has two weaknesses. First, as typical natural signals, both visual and audio features have heavy spatial-temporal redundancy and noisy information that are useless for other modalities. Second, the computation of cross-modal attention is costly, with square complexity when calculating clip-to-clip correlations. A recent work \cite{nagrani2021attention} tried to tackle the first problem by introducing attention bottlenecks that can be regarded as the information bridge across modalities. Although promising results have been achieved, this module still suffers from high computational cost since inter- and cross-modal correlations are jointly modeled. In this work, we extend this idea and propose to disentangle these strategies, thus the resulting bottleneck transformer module can be divided into two stages, \ie feature compression and expansion, shown in Figure~\ref{fig:3}.

\paragraph{Feature Compression}

Following \cite{nagrani2021attention}, we introduce bottleneck tokens $\{z_i\}_{i=1}^{N_b}$ to capture the compressed features from all modalities. Here $N_b$ is a number much smaller than the number of video clips $N_v$. The feature compression is realized by several multi-head attentions between bottleneck tokens and the features from different modalities. Since there are only visual and audio modalities in this case, the compression process can be represented by
\begin{gather}\label{eq:2}
z_i' = z_i + w_z \sum_{j=1}^{N_v} \frac{\exp(w_q z_i \times w_k x_j)}{\sum_{m=1}^{N_v} \exp(w_q z_i \times w_k x_m)} w_v x_j
\end{gather}
where $z_i$ and $z_i'$ are input and output features of bottleneck tokens. Other notations are consistent with Eq.~\ref{eq:1}. The only difference between Eq.~\ref{eq:1} and Eq.~\ref{eq:2} is that the query matrix is replaced by $z_i$, aiming to aggregate features into bottleneck tokens. We apply this operation for both visual and audio features, so that multi-modality information is refined and compressed into bottleneck tokens.

\paragraph{Feature Expansion}

After compressing the multi-modal information, we expand the features and propagate them into each modality using another multi-head attention. Formally, the computation is as follows.
\begin{gather}
x_i' = x_i + w_z \sum_{j=1}^{N_v} \frac{\exp(w_q x_i \times w_k z_j)}{\sum_{m=1}^{N_v} \exp(w_q x_i \times w_k z_m)} w_v z_j
\end{gather}
Here, $x_i'$ represents the cross-modality enhanced features of clip $i$. These features are then fed into feed-forward networks for further projection. Leveraging such a two-stage feature propagation across modalities, visual and audio features are augmented under linear computational complexities, without incorporating noisy information.

\subsection{Query Generator}

As transformers are firstly introduced for language translation tasks, the lengths of the input and output sequences may not be the same, where the length of the output sequence is determined by the query embeddings fed into the decoder. When generalized to vision tasks, query embeddings are randomly initialized and learned during training. Such a scheme may not be suitable for video highlight detection, since the outputs ought to be strictly aligned with the input tokens. Besides, query embeddings shall naturally guide the process of representation decoding. Therefore, we introduce a query generator to adaptively generate temporally aligned moment queries depending on the natural language input. This module is also constructed by a multi-head attention layer, in which visual-audio joint representations $\{r_i\}_{i=1}^{N_v}$ act as query, textual features are key and value. Our hypothesis is that by computing the attention weights between video clips and text queries, each clip can learn whether it contains which of the concepts described in the text, and predict a query embedding that can be used to decode the learned information for different needs. Note that when text queries are not available, joint representations and learnable positional encodings are summed up to serve as moment queries instead.

\begin{table}
\footnotesize
\setlength{\tabcolsep}{0pt}
\caption{Experimental results on QVHighlights \texttt{test} split. MR and HD represent moment retrieval and highlight detection, respectively. w/ PT means pre-training with ASR captions.}
\label{tab:1}
\begin{tabularx}{\linewidth}{@{\hspace{1mm}}p{2.65cm}p{0.765cm}<{\centering}p{0.765cm}<{\centering}p{0.8mm}<{\centering}p{0.765cm}<{\centering}p{0.765cm}<{\centering}p{0.765cm}<{\centering}p{0.8mm}<{\centering}p{0.7cm}<{\centering}p{0.875cm}<{\centering}}
\toprule
& \multicolumn{6}{c}{\textbf{MR}} & & \multicolumn{2}{c}{\textbf{HD}} \\
\cmidrule{2-7} \cmidrule{9-10}
& \multicolumn{2}{c}{R1} & & \multicolumn{3}{c}{mAP} & & \multicolumn{2}{c}{$\geq$ Very Good} \\
\cmidrule{2-3} \cmidrule{5-7} \cmidrule{9-10}
\vspace{-0.73cm}\hspace{0.8cm}\textbf{Method} & @0.5 & @0.7 & & @0.5 & @0.75 & Avg. & & mAP & HIT@1 \\
\midrule
BeautyThumb \cite{song2016click} & -- & -- & & -- & -- & -- & & 14.36 & 20.88 \\
DVSE \cite{liu2015multi} & -- & -- & & -- & -- & -- & & 18.75 & 21.79 \\
MCN \cite{anne2017localizing} & 11.41 & 2.72 & & 24.94 & 8.22 & 10.67 & & -- & -- \\
CAL \cite{escorcia2019temporal} & 25.49 & 11.54 & & 23.40 & 7.65 & 9.89 & & -- & -- \\
XML \cite{lei2020tvr} & 41.83 & 30.35 & & 44.63 & 31.73 & 32.14 & & 34.49 & 55.25 \\
XML+ \cite{lei2021qvhighlights} & 46.69 & 33.46 & & 47.89 & 34.67 & 34.90 & & 35.38 & 55.06 \\
\midrule
Moment-DETR \cite{lei2021qvhighlights} & 52.89 & 33.02 & & 54.82 & 29.40 & 30.73 & & 35.69 & 55.60 \\
Moment-DETR w/ PT & 59.78 & 40.33 & & \textbf{60.51} & 35.36 & 36.14 & & 37.43 & 60.17 \\
\midrule
\textbf{UMT} (Ours) & 56.23 & 41.18 & & 53.83 & 37.01 & 36.12 & & 38.18 & 59.99 \\
\textbf{UMT} (Ours) w/ PT & \textbf{60.83} & \textbf{43.26} & & 57.33 & \textbf{39.12} & \textbf{38.08} & & \textbf{39.12} & \textbf{62.39} \\
\bottomrule
\end{tabularx}
\end{table}

\subsection{Query Decoder and Prediction Heads}

Query decoder takes visual-audio joint representations $\{r_i\}_{i=1}^{N_v}$ and text-guided moment queries $\{q_i\}_{i=1}^{N_v}$ as inputs, and decodes the video features for joint moment retrieval and highlight detection. The output sequence of the query decoder has the same length as the encoder input. Such a peculiarity has two advantages: 1) We may obtain the clip-level saliency (highlight) scores as simple as adding a linear projection layer with sigmoid activation. 2) The dynamic length of output sequence also enables us to define moment retrieval as a \textit{keypoint detection} problem \cite{law2018cornernet,zhou2019objects}. That is, each moment can be represented by its temporal center and duration (window), where the center point can be estimated by predicting a temporal heatmap and extracting local maxima. The window can be further regressed from features of the center. Note that the points in the heatmap are discrete, which may be misaligned with the real temporal center and inevitably damage the retrieval performances. An extra offset term used to adjust the center should also be predicted. We adopt four linear projection layers to predict the saliencies, centers, windows, and offsets, respectively.

\begin{table}
\footnotesize
\setlength{\tabcolsep}{0pt}
\caption{Comparison with representative moment retrieval methods on Charades-STA \texttt{test} split. All the models use the officially released VGG and/or optical flow features of Charades.}
\label{tab:2}
\begin{threeparttable}
\begin{tabularx}{\linewidth}{@{\hspace{2mm}}p{1.9cm}p{1.5cm}<{\centering}p{1.5cm}<{\centering}p{1.5mm}<{\centering}p{1.5cm}<{\centering}p{1.5cm}<{\centering}}
\toprule
\multirow{2}{*}{\vspace{-0.2cm}\hspace{0.4cm}\textbf{Method}} & \multicolumn{2}{c}{R@1} & & \multicolumn{2}{c}{R@5} \\
\cmidrule{2-3} \cmidrule{5-6}
& IoU=0.5 & IoU=0.7 & & IoU=0.5 & IoU=0.7 \\
\midrule
SAP \cite{chen2019semantic} & 27.42 & 13.36 & & 66.37 & 38.15 \\
SM-RL \cite{wang2019language} & 24.36 & 11.17 & & 61.25 & 32.08 \\
MAN \cite{zhang2019man} & 41.24 & 20.54 & & 83.21 & 51.85 \\
2D-TAN \cite{zhang2020learning} & 40.94 & 22.85 & & 83.84 & 50.35 \\
FVMR \cite{gao2021fast} & 42.36 & 24.14 & & 83.97 & 50.15 \\
\midrule
\textbf{UMT}\tnote{\dag} (Ours) & 48.31 & \textbf{29.25} & & 88.79 & \textbf{56.08} \\
\textbf{UMT}\tnote{\ddag} (Ours) & \textbf{49.35} & 26.16 & & \textbf{89.41} & 54.95 \\
\bottomrule
\end{tabularx}
\vspace{0.5mm}
\hspace{0.3mm}
\tnote{\dag}\,video\,+\,audio,\, \tnote{\ddag}\,video\,+\,optical\,flow
\end{threeparttable}
\end{table}

\begin{table*}
\begin{floatrow}
\renewcommand\tabcolsep{0pt}
\footnotesize
\hspace{-0.175cm}
\ttabbox[0.405\textwidth]{
\caption{Experimental results on YouTube Highlights (metric: mAP). Above are the methods using visual features only, the others are using visual-audio features.}}{
\label{tab:3}
\begin{tabularx}{\linewidth}{@{\hspace{1mm}}p{2cm}|@{\hspace{0.5mm}}p{0.7cm}<{\centering}p{0.7cm}<{\centering}p{0.7cm}<{\centering}p{0.7cm}<{\centering}p{0.7cm}<{\centering}p{0.7cm}<{\centering}p{0.7cm}<{\centering}}
\toprule
\textbf{Method} & \textbf{Dog} & \textbf{Gym.} & \textbf{Par.} & \textbf{Ska.} & \textbf{Ski.} & \textbf{Sur.} & \textbf{Avg.} \\
\midrule
RRAE \cite{yang2015unsupervised} & 49.0 & 35.0 & 50.0 & 25.0 & 22.0 & 49.0 & 38.3 \\
GIFs \cite{gygli2016video2gif} & 30.8 & 33.5 & 54.0 & 55.4 & 32.8 & 54.1 & 46.4 \\
LSVM \cite{sun2014ranking} & 60.0 & 41.0 & 61.0 & 62.0 & 36.0 & 61.0 & 53.6 \\
LIM-S \cite{xiong2019less} & 57.9 & 41.7 & 67.0 & 57.8 & 48.6 & 65.1 & 56.4 \\
SL-Module \cite{xu2021cross} & \textbf{70.8} & 53.2 & 77.2 & \textbf{72.5} & 66.1 &76.2 & 69.3 \\
\midrule
MINI-Net \cite{hong2020mini} & 58.2 & 61.7 & 70.2 & 72.2 & 58.7 & 65.1 & 64.4 \\
TCG \cite{ye2021temporal} & 55.4 & 62.7 & 70.9 & 69.1 & 60.1 & 59.8 & 63.0 \\
Joint-VA \cite{badamdorj2021joint} & 64.5 & 71.9 & 80.8 & 62.0 & \textbf{73.2} & 78.3 & 71.8 \\
\textbf{UMT} (Ours) & 65.9 & \textbf{75.2} & \textbf{81.6} & 71.8 & 72.3 & \textbf{82.7} & \textbf{74.9} \\
\bottomrule
\end{tabularx}}
\hspace{-0.5cm}
\ttabbox[0.565\textwidth]{
\caption{Comparison with representative highlight detection methods on TVSum (metric: Top-5 mAP). Above are the methods using visual features only, the others are using visual-audio features.}}{
\label{tab:4}
\begin{tabularx}{\linewidth}{@{\hspace{1mm}}p{2cm}|@{\hspace{0.5mm}}p{0.7cm}<{\centering}p{0.7cm}<{\centering}p{0.7cm}<{\centering}p{0.7cm}<{\centering}p{0.7cm}<{\centering}p{0.7cm}<{\centering}p{0.7cm}<{\centering}p{0.7cm}<{\centering}p{0.7cm}<{\centering}p{0.7cm}<{\centering}p{0.7cm}<{\centering}}
\toprule
\textbf{Method} & \textbf{VT} & \textbf{VU} & \textbf{GA} & \textbf{MS} & \textbf{PK} & \textbf{PR} & \textbf{FM} & \textbf{BK} & \textbf{BT} & \textbf{DS} & \textbf{Avg.} \\
\midrule
sLSTM \cite{zhang2016video} & 41.1 & 46.2 & 46.3 & 47.7 & 44.8 & 46.1 & 45.2 & 40.6 & 47.1 & 45.5 & 45.1 \\
SG \cite{mahasseni2017unsupervised} & 42.3 & 47.2 & 47.5 & 48.9 & 45.6 & 47.3 & 46.4 & 41.7 & 48.3 & 46.6 & 46.2 \\
LIM-S \cite{xiong2019less} & 55.9 & 42.9 & 61.2 & 54.0 & 60.4 & 47.5 & 43.2 & 66.3 & 69.1 & 62.6 & 56.3 \\
Trailer \cite{wang2020trailer} & 61.3 & 54.6 & 65.7 & 60.8 & 59.1 & 70.1 & 58.2 & 64.7 & 65.6 & 68.1 & 62.8 \\
SL-Module \cite{xu2021cross} & 86.5 & 68.7 & 74.9 & \textbf{86.2} & 79.0 & 63.2 & 58.9 & 72.6 & 78.9 & 64.0 & 73.3 \\
\midrule
MINI-Net \cite{hong2020mini} & 80.6 & 68.3 & 78.2 & 81.8 & 78.1 & 65.8 & 57.8 & 75.0 & 80.2 & 65.5 & 73.2 \\
TCG \cite{ye2021temporal} & 85.0 & 71.4 & 81.9 & 78.6 & 80.2 & 75.5 & 71.6 & 77.3 & 78.6 & 68.1 & 76.8 \\
Joint-VA \cite{badamdorj2021joint} & 83.7 & 57.3 & 78.5 & 86.1 & 80.1 & 69.2 & 70.0 & 73.0 & \textbf{97.4} & 67.5 & 76.3 \\
\textbf{UMT} (Ours) & \textbf{87.5} & \textbf{81.5} & \textbf{88.2} & 78.8 & \textbf{81.4} & \textbf{87.0} & \textbf{76.0} & \textbf{86.9} & 84.4 & \textbf{79.6} & \textbf{83.1} \\
\bottomrule
\end{tabularx}}
\end{floatrow}
\end{table*}

During training, the clip-level saliency score prediction is optimized using a binary cross-entropy loss $\mathcal{L}_s$. For each ground truth moment with center $p \in [1, N_v]$ and window $d$, we quantize the center point to $\widetilde{p}$ and fill the heatmap $H \in [0, 1]^{N_v}$ using a 1D gaussian kernel $H_x = \exp(-\frac{(x - \widetilde{p})^2}{2{\sigma_p}^2})$, where $x$ is the temporal coordinate and $\sigma_p$ is the window-adaptive standard deviation. We optimize the center point prediction using the gaussian focal loss \cite{lin2017focal} as
\begin{gather}
\mathcal{L}_c = -\frac{1}{N} \sum_{x}
\begin{cases}
(1 - \hat{H_x})^\alpha \log(\hat{H_x}) & if H_x = 1 \\
(1 - H_x)^\gamma \hat{H_x}^\alpha \log(1 - \hat{H_x}) & otherwise
\end{cases}
\end{gather}
Here, $N$ is the number of moments, $\alpha$ and $\gamma$ denote the weighting and the exponent of the modulating factors in the focal loss, which are set to 2.0 and 4.0 in practice. For window and offset regression, we simply adopt L1 losses to optimize the actual values for all ground truth centers as
\begin{gather}
\mathcal{L}_w = -\frac{1}{N} \sum_{p} |w_p - \hat{w_p}| \\
\mathcal{L}_o = -\frac{1}{N} \sum_{p} |(o_p - \widetilde{p}) - \hat{o_p}|
\end{gather}
where $w_p$, $\hat{w_p}$, $o_p$, and $\hat{o_p}$ are the ground truth and predicted windows and offsets respectively. The overall training loss would be the weighted sum of all the losses above as
\begin{gather}
\mathcal{L} = \lambda_s \mathcal{L}_s + \lambda_c \mathcal{L}_c + \lambda_w \mathcal{L}_w + \lambda_o \mathcal{L}_o
\end{gather}
where $\lambda_{\{s, c, w, o\}}$ are the weights for saliency, center, window, and offset losses, respectively. When testing, the moment boundries are obtained by comining the center, window, and offset terms as introduced in \cite{law2018cornernet,zhou2019objects}.

\section{Experiments}

\subsection{Datasets and Experimental Settings}

\paragraph{Datasets}

QVHighlights \cite{lei2021qvhighlights} is the only existing public dataset that has ground-truth annotations for both moment retrieval and highlight detection, thus being suitable for evaluating the full version of our proposed model. This dataset contains videos cropped into 10,148 short (150s-long) segments, and had each segment annotated with at least one text query depicting its relevant moments. There are averagely about 1.8 disjoint moments per query, annotated on non-overlapping 2s-long clips. In total, there are 10,310 queries with 18,367 annotated moments. We follow the original QVHighlights data splits in all experiments.

We also utilize three more datasets: Charades-STA \cite{gao2017tall}, YouTube Highlights \cite{sun2014ranking}, and TVSum \cite{song2015tvsum} for further evaluation on the moment retrieval or highlight detection task only, as our model has the flexibility for tasks. Charades-STA contains 16,128 query-moment pairs annotating different actions. YouTube Highlights has 6 domains with 433 videos currently available. TVSum includes 10 domains with 5 videos each. We follow the tradition to work on a random 0.8/0.2 training/test split. Note that the annotators for TVSum were aware of the video titles, so we believe that these titles can serve as noisy text queries. Our model's flexibility to handle this situation is studied.

\paragraph{Evaluation Metrics}

We use the same evaluation metrics used in existing works. Specifically, for QVHighlights, Recall@1 with IoU thresholds 0.5 and 0.7, mean average precision (mAP) with IoU thresholds 0.5 and 0.75, and the average mAP over a series of IoU thresholds [0.5:0.05:0.95] are used for moment retrieval. For highlight detection, mAP and HIT@1 are utilized, where a clip prediction is treated as a true positive if it has the saliency score of \emph{Very Good}. For Charades-STA, Recall@1 and Recall@5 with IoU thresholds 0.5 and 0.7 are used. For YouTube Highlights and TVSum, mAP and Top-5 mAP are adopted, respectively.

\paragraph{Implementation Details}

On QVHighlights, we simply leverage the pre-extracted features using SlowFast \cite{feichtenhofer2019slowfast} and CLIP \cite{radford2021learning}. Official VGG \cite{simonyan2015very} and optical flow features, as well as GloVe \cite{pennington2014glove} embeddings, are used for Charades-STA. On YouTube Highlights and TVSum, we obtain clip-level visual features using an I3D \cite{carreira2017quo} pre-trained on Kinetics 400 \cite{kay2017kinetics}. Since each feature vector captures 32 consecutive frames, we follow \cite{badamdorj2021joint} and consider the feature vector belonging to a clip if their overlap is more than 50\%. We also use CLIP to extract the title features in TVSum. Audio features of all the datasets are extracted by a PANN \cite{kong2020panns} model pre-trained on AudioSet \cite{gemmeke2017audio}. Visual and audio features are temporally aligned at clip level.

\begin{figure*}
\centering
\includegraphics[width=\linewidth]{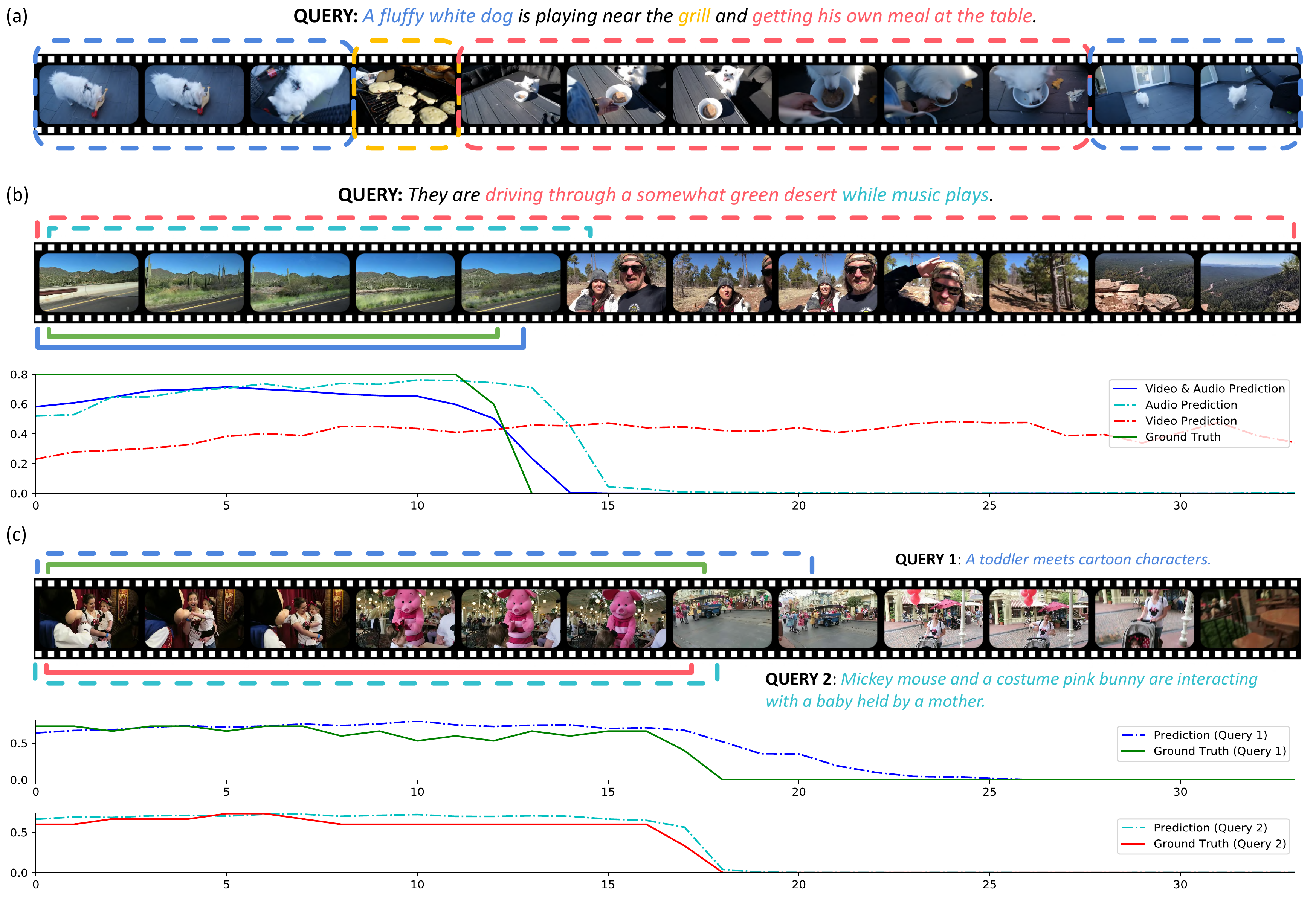}
\caption{Qualitative results on QVHighlights. The predicted moments and saliency scores are shown by brackets and lines. a) All the highlight clips are presented, indicating that UMT can learn implicit correlations between video content and query semantics. b) Different modality combinations guide the model pay attention to different moments. c) Our model can handle multiple queries in a single video.}
\label{fig:4}
\end{figure*}

All the models in our experiments contain one uni-modal and cross-modal encoder layer each. The number of decoder layers is set to 3 for QVHighlights and Charades-STA, and 1 for YouTube Highlights and TVSum since they have smaller scales. The number of bottleneck tokens $N_b$ is insensitive thus be set to 4. The weights of losses are set as $\lambda_s = 3.0$, $\lambda_c = 1.0$, $\lambda_w = 0.1$, and $\lambda_o = 1.0$, while $\lambda_w$ and $\lambda_o$ are reduced to 0.05 and 0.5 specially for Charades-STA. Following \cite{lei2021qvhighlights}, we set the hidden dimensions to 256, with $4\times$ dimension expansions in feed forward networks. Learnable positional encodings, pre-norm style layer normalizations \cite{ba2016layer}, 8 attention heads, and 0.1 dropout rates are used in all transformer layers. We also adopt extra pre-dropouts with rate 0.5 for visual and audio inputs, and 0.3 for text inputs. In all experiments, we use Adam \cite{kingma2015adam} optimizer with 1e-3 learning rate and 1e-4 weight decay. The model is trained with batch size 32 for 200 epochs on QVHighlights, batch size 8 for 100 epochs on Charades-STA, batch size 4 for 100 epochs on YouTube Highlights, and batch size 1 for 500 epochs on TVSum, respectively.

\subsection{Results on Joint Video Moment Retrieval and Highlight Detection}

We first evaluate our proposed UMT on QVHighlights \texttt{test} split. The results are shown in Table~\ref{tab:1}, in comparison with all the other performances ever reported. On both moment retrieval and highlight detection tasks, our proposed model outperforms all the existing approaches, including the previous state-of-the-art method Moment-DETR \cite{lei2021qvhighlights} under both settings (with or without pre-training with automatic speech recognition captions). Figure~\ref{fig:4} presents some qualitative results of our method on QVHighlights.

\subsection{Results on Moment Retrieval}

Table~\ref{tab:2} shows the comparison of UMT with some representative methods on Charades-STA \texttt{test} split. Our approach performs better than previous methods under different metrics. We also tried adopting optical flow instead of audio and obtained similar performances.

\subsection{Results on Highlight Detection}

Highlight detection results on YouTube Highlights and TVSum are presented in Table~\ref{tab:2} and Table~\ref{tab:3}, respectively. On both datasets, UMT performs better than not only representative methods which only use video, but also existing multi-modal ones that utilize both video and audio. 

\subsection{Ablation Studies}

\paragraph{Multi-modality (Visual-Audio)}

Table~\ref{tab:5} shows the performances of all the multi-modal methods when different modalities are used. Note that Moment-DETR$^+$ is a multi-modal extension of original Moment-DETR \cite{lei2021qvhighlights} by implementing a similar bottleneck structure as UMT. Clearly, multi-modal learning can significantly boost most methods' performance on all the datasets and tasks in comparison with using a single modality, since it can capture more useful information. Compared with the most similar competitor Moment-DETR$^+$, UMT can better explore the complementary information from different modalities and suppress the possible noise during information transfer.

\begin{table}
\footnotesize
\setlength\tabcolsep{0pt}
\caption{The effectiveness of multi-modal learning on YouTube Highlights, TVSum, and QVHighlights \texttt{val} split. MR and HD denote moment retrieval and highlight detection, respectively.}
\label{tab:5}
\begin{threeparttable}
\begin{tabularx}{\linewidth}{@{\hspace{1.2mm}}p{2.4cm}p{1.4cm}<{\centering}p{1mm}<{\centering}p{1.4cm}<{\centering}p{1mm}<{\centering}p{1.4cm}<{\centering}p{1.4cm}<{\centering}}
\toprule
& \textbf{YouTube} & & \textbf{TVSum} & & \multicolumn{2}{c}{\textbf{QVHighlights}} \\
\cmidrule{2-2}\cmidrule{4-4}\cmidrule{6-7}
\vspace{-0.48cm}\hspace{0.75cm}\textbf{Method} & mAP & & Top-5 mAP & & MR (mAP) & HD (mAP) \\
\midrule
MINI-Net\tnote{\dag} \cite{hong2020mini} & 61.38 & & 69.79 & & -- & -- \\
Joint-VA\tnote{\dag} \cite{badamdorj2021joint} & 70.50 & & 74.80 & & -- & -- \\
\makebox[0pt][l]{Moment-DETR\tnote{\dag} \cite{lei2021qvhighlights}} & -- & & -- & & 32.20 & 36.52 \\
\textbf{UMT}\tnote{\dag} (Ours) & 73.48 & & 81.89 & & 37.79 & 38.97 \\
\midrule
MINI-Net\tnote{\ddag} \cite{hong2020mini} & 52.23 & & 59.72 & & -- & -- \\
Joint-VA\tnote{\ddag} \cite{badamdorj2021joint} & 67.00 & & 68.70 & & -- & -- \\
\makebox[0pt][l]{Moment-DETR\tnote{\ddag} \cite{lei2021qvhighlights}} & -- & & -- & & 16.69 & 26.00 \\
\textbf{UMT}\tnote{\ddag} (Ours) & 65.61 & & 76.51 & & 13.73 & 23.91 \\
\midrule
MINI-Net \cite{hong2020mini} & 64.36 & & 73.24 & & -- & -- \\
Joint-VA \cite{badamdorj2021joint} & 71.80 & & 76.30 & & -- & -- \\
\makebox[0pt][l]{Moment-DETR\tnote{$+$} \cite{lei2021qvhighlights}} & -- & & -- & & 34.05 & 37.67 \\
\textbf{UMT} (Ours) & \textbf{74.93} & & \textbf{83.14} & & \textbf{38.59} & \textbf{39.85} \\
\bottomrule
\end{tabularx}
\vspace{0.5mm}
\hspace{0.3mm}
\tnote{\dag}\,video\,only,\, \tnote{\ddag}\,audio\,only,\, \tnote{$+$}\,w/\,bottleneck\,transformer
\end{threeparttable}
\end{table}

\begin{table}
\footnotesize
\setlength{\tabcolsep}{0pt}
\caption{Comparison with Moment-DETR using different training task combinations on QVHighlights \texttt{val} split. MR and HD denote moment retrieval and highlight detection, respectively.}
\label{tab:6}
\begin{tabularx}{\linewidth}{@{\hspace{1mm}}p{1.55cm}<{\centering}p{0.8cm}<{\centering}p{0.8cm}<{\centering}p{1mm}<{\centering}p{0.95cm}<{\centering}p{0.95cm}<{\centering}p{0.95cm}<{\centering}p{1mm}<{\centering}p{0.95cm}<{\centering}p{0.95cm}<{\centering}}
\toprule
\multirow{3}{*}{\vspace{-0.2cm}\textbf{Method}} & \multicolumn{2}{c}{\textbf{Tr. Task(s)}} & & \multicolumn{3}{c}{\textbf{MR}} & & \multicolumn{2}{c}{\textbf{HD}} \\
\cmidrule{2-3} \cmidrule{5-7} \cmidrule{9-10}
& \multirow{2}{*}{MR} & \multirow{2}{*}{HD} & & R1 & R1 & mAP & & \multirow{2}{*}{mAP} & \multirow{2}{*}{HIT@1} \\
& & & & @0.5 & @0.7 & Avg. & & & \\
\midrule
\multirow{2}{*}{Moment-} & \checkmark & & & 44.84 & 25.87 & 25.05 & & -- & -- \\
\multirow{2}{*}{DETR \cite{lei2021qvhighlights}} & & \checkmark & & -- & -- & -- & & 36.52 & 56.45 \\
& \checkmark & \checkmark & & 53.94 & 34.84 & 32.20 & & 35.65 & 55.55 \\
\midrule
\multirow{2}{*}{{\textbf{UMT}}} & \checkmark & & & 54.14 & 33.82 & 34.02 & & -- & -- \\
\multirow{2}{*}{{(Ours)}} & & \checkmark & & -- & -- & -- & & \textbf{40.22} & \textbf{65.03} \\
& \checkmark & \checkmark & & \textbf{60.26} & \textbf{44.26} & \textbf{38.59} & & 39.85 & 64.19 \\
\bottomrule
\end{tabularx}
\end{table}

\paragraph{Multi-task Co-optimization}

Given a text query for a video, retrieving the related moments and detecting salient highlights in such moments seem to be highly correlated tasks. Therefore, it is interesting to see how the multi-task co-optimization performs in comparison with training for each individual task when the same framework and backbone are used. We conduct single-task experiments by turning off the losses corresponding to each task and training the rest of the model. Note that moment retrieval is considered to be a harder task than highlight detection as explained in Section~\ref{sec:1}. The results in Table~\ref{tab:6} show that the co-optimization not only generates results for both tasks simultaneously, but also significantly boosts the performance on moment retrieval. This is clear for both our UMT and Moment-DETR \cite{lei2021qvhighlights}. Note that when training for moment retrieval only, UMT performs much better than Moment-DETR, indicating its superiority on the model design. Our UMT better models the moment retrieval task as a keypoint detection problem \cite{law2018cornernet,zhou2019objects} rather than set prediction or clip classification. Moreover, the inputs to the UMT decoder are clip-aligned text-guided queries instead of positional encodings, which enables more flexible output sequence lengths and may provide stronger query information for each clip. Therefore, we believe that UMT can model the relationship between the two tasks better than Moment-DETR does.

\paragraph{Moment Retrieval Losses}

Table~\ref{tab:7} presents the performances of UMT when different combinations of moment retrieval losses are used. Since the center loss $\mathcal{L}_{c}$ and the window loss $\mathcal{L}_{w}$ are mandatory for representing a moment, only the necessity of the offset loss $\mathcal{L}_{o}$ is justified. As the comparison shows, modeling the temporal offset does make the moment boundary prediction more accurate.

\paragraph{Justification of Text Queries}

We believe that highlight detection based on text queries is an important setting for highlight detection, as different interests can favor quite different highlights from the same video. Table~\ref{tab:8} reports the results of our model, with or without the text queries. It can be seen that when the query is relevant, it does improve the highlight detection performance, and such an improvement is more significant when the relevance is greater.

\begin{table}
\footnotesize
\setlength\tabcolsep{0pt}
\caption{Effectiveness justification of the offset loss for moment retrieval on QVHighlights \texttt{val} split. Both models are trained using the co-optimization recipe.}
\label{tab:7}
\begin{tabularx}{0.825\linewidth}{@{\hspace{1.5mm}}p{1.9cm}<{\centering}p{1mm}<{\centering}p{0.9cm}<{\centering}p{0.9cm}<{\centering}p{1mm}<{\centering}p{0.9cm}<{\centering}p{0.9cm}<{\centering}p{0.9cm}<{\centering}}
\toprule
\multirow{3}{*}{\vspace{0.15cm}\textbf{Losses}} & & \multicolumn{2}{c}{R1} & & \multicolumn{3}{c}{mAP} \\
\cmidrule{3-4}\cmidrule{6-8}
& & @0.5 & @0.7 & & @0.5 & @0.75 & Avg. \\
\midrule
$\mathcal{L}_{c}$ + $\mathcal{L}_{w}$ & & \textbf{62.32} & 43.23 & & \textbf{57.78} & 38.61 & 37.36 \\
$\mathcal{L}_{c}$ + $\mathcal{L}_{w}$ + $\mathcal{L}_{o}$ & & 60.26 & \textbf{44.26} & & 56.70 & \textbf{39.90} & \textbf{38.59} \\
\bottomrule
\end{tabularx}
\end{table}

\begin{table}
\footnotesize
\setlength\tabcolsep{0pt}
\caption{The influence of weakly relevant (TVSum) or highly relevant (QVHighlights) text queries on highlight detection.}
\label{tab:8}
\begin{tabularx}{0.66\linewidth}{@{\hspace{2mm}}p{1.4cm}<{\centering}p{1.4cm}<{\centering}p{1.2cm}<{\centering}p{1.2cm}<{\centering}}
\toprule
\multirow{2}{*}{\vspace{-0.15cm}\textbf{Text Query}} & \multirow{2}{*}{\vspace{-0.15cm}\textbf{TVSum}} & \multicolumn{2}{c}{\textbf{QVHighlights}} \\
\cmidrule{3-4}
& & mAP & HIT@1 \\
\midrule
& 81.42 & 25.14 & 33.42 \\
\checkmark & \textbf{83.14} & \textbf{39.85} & \textbf{64.19} \\
\bottomrule
\end{tabularx}
\end{table}

\section{Conclusion}

This paper introduces a novel and also the first framework for solving joint moment retrieval and highlight detection as well as its individual component problems in a unified way. It is also the first to integrate multi-modal learning into its model for such a purpose. The effectiveness and superiority of the proposal have been demonstrated on diverse and representative public datasets, in comparison with relevant methods under various settings. The framework is robust to modality quality variations and also flexible enough to work under different text query conditions.

\section*{Acknowledgements}

This research is supported in part by Key-Area Research and Development Program of Guangdong Province, China with Grant 2019B010155002 and financial support from ARC Lab, Tencent PCG.

\small
\bibliographystyle{cvpr}
\bibliography{main}

\iftoggle{cvprpagenumbers}{
\vfill
\normalsize
\setcounter{section}{0}
\renewcommand{\thesection}{\Alph{section}}
\setcounter{equation}{0}
\renewcommand{\theequation}{\Alph{equation}}
\setcounter{table}{0}
\renewcommand{\thetable}{\Alph{table}}
\setcounter{figure}{0}
\renewcommand{\thefigure}{\Alph{figure}}

\section*{\Large Appendix}
\vspace{0.5mm}

In this document, we provide more descriptions of the model architecture and implementation details to complement the main paper. Additional ablation studies and visualization on QVHighlights \cite{lei2021qvhighlights} are also incorporated to demonstrate the effectiveness of the proposed method.

\section{Model Architecture}

Learnable positional encodings with $0.1$ dropout rates are adopted in all the encoder and decoder layers. More specifically, in uni-modal encoders, the same positional encodings are added to the $Q$ and $K$ matrices. In cross-modal encoders, they are added to the $K$ matrix during feature compression and the $Q$ matrix during feature expansion. In the query decoder, two independent positional encodings are added to the $Q$ and $K$ matrices, respectively.

In video- or audio-only schemes, cross-modal encoders are not necessary thus be removed. Their output normalization layers are moved to the end of the corresponding uni-modal encoders. When text queries are not available, the query generator simply outputs the visual-audio joint representations $\{r_i\}_{i=1}^{N_v}$, which will be further added with learnable positional encodings to construct moment queries.

\section{Implementation Details}

Similar to \cite{law2018cornernet,zhou2019objects}, during training, each ground truth moment with quantized center $\widetilde{p}$ and window $d$ is used to establish a 1D gaussian kernel $H_x = \exp(-\frac{(x - \widetilde{p})^2}{2{\sigma_p}^2})$ with radius $r_p$ on the heatmap $H$. Here, $x$ indicates the temporal coordinate aligned with clip indices, $r_p$ and $\sigma_p$ are window-adaptive parameters that can be computed as
\begin{gather}
r_p = \mu \cdot d \\
\sigma_p = \rho \cdot (r_p + 1) \label{eq:b}
\end{gather}
where $\mu$ and $\rho$ are hyperparameters controlling the corresponding values. We add $1$ to $r_p$ in Eq.~\ref{eq:b} to ensure that the output $\sigma_p$ is not too small, preventing extremely large values in the heatmap. We observe that the moment retrieval performance is not sensitive to these hyperparameters, thus both of them are set to $0.2$ in all experiments. When testing, we compute the moment retrieval results assuming all the clips are centers to obtain a higher recall.

Following \cite{lei2021qvhighlights}, we also consider the weakly-supervised pre-training on QVHighlights with automatic speech recognition (ASR) captions. During pre-training, the saliency loss is turned off since the supervision is only for moment retrieval. We observe that UMT converges much faster than Moment-DETR due to the novel formulation of moment retrieval, thus we increase the batch size to 2048 and pre-train the model for 100 epochs to prevent overfitting. Other hyperparameters strictly follow the original settings.

\vfill
\vspace{0.25mm}

\begin{table}
\footnotesize
\setlength\tabcolsep{0pt}
\caption{Comparison of cross-modal fusion methods on YouTube Highlights, TVSum, and QVHighlights \texttt{val} split. MR and HD denote moment retrieval and highlight detection, respectively.}
\label{tab:a}
\begin{tabularx}{\linewidth}{p{1.75cm}<{\centering}p{1.55cm}<{\centering}p{1mm}<{\centering}p{1.55cm}<{\centering}p{1mm}<{\centering}p{1.55cm}<{\centering}p{1.55cm}<{\centering}}
\toprule
& \textbf{YouTube} & & \textbf{TVSum} & & \multicolumn{2}{c}{\textbf{QVHighlights}} \\
\cmidrule{2-2}\cmidrule{4-4}\cmidrule{6-7}
\vspace{-0.48cm}\textbf{Method} & mAP & & Top-5 mAP & & MR (mAP) & HD (mAP) \\
\midrule
Concat & 73.32 & & 80.26 & & 37.03 & 38.74 \\
Mean & 73.29 & & 81.76 & & 37.04 & 38.91 \\
Sum & 73.53 & & 81.77 & & 37.33 & 38.88 \\
Bottleneck & \textbf{74.93} & & \textbf{83.14} & & \textbf{38.59} & \textbf{39.85} \\
\bottomrule
\end{tabularx}
\end{table}

\begin{table}
\footnotesize
\renewcommand\tabcolsep{2pt}
\caption{Ablation on number of tokens in the bottleneck transformer on QVHighlights \texttt{val} split (metric: mAP). MR and HD denote moment retrieval and highlight detection, respectively.}
\label{tab:b}
\begin{tabularx}{\linewidth}{p{1.75cm}<{\centering}|p{0.9cm}<{\centering}p{0.9cm}<{\centering}p{0.9cm}<{\centering}p{0.9cm}<{\centering}p{0.9cm}<{\centering}p{0.9cm}<{\centering}}
\toprule
\textbf{\#Tokens} & \textbf{4} & \textbf{8} & \textbf{16} & \textbf{32} & \textbf{64} & \textbf{128} \\
\midrule
MR & \textbf{38.59} & 37.19 & 38.29 & 37.15 & 37.20 & 38.18 \\
HD & \textbf{39.85} & 39.64 & 39.26 & 39.37 & 39.22 & 39.26 \\
\bottomrule
\end{tabularx}
\end{table}

\section{Ablation Studies}

Table~\ref{tab:a} shows the comparison among bottleneck tokens and its baselines for cross-modal feature fusion on multiple datasets. It shows that utilizing the bottleneck transformer rather than simple operations can improve the performance on both moment retrieval and highlight detection.

Table~\ref{tab:b} studies the influence of the number of bottleneck tokens. We observe that the performance is insensitive to the number of tokens, since the feature compression and expansion process eliminate the undesirable noise.

\section{Visualization}

Figure~\ref{fig:a} displays more qualitative results on QVHighlights \cite{lei2021qvhighlights}. The results show that visual and audio features contribute to different moment retrieval and highlight detection outcomes. For example, in Figure~\ref{fig:a}~(b), the video-only model fails to refine the retrieved moment given the determiner `indicating we are listening to his audio', while the audio-only model can not distinguish the moment boundaries. Combining the visual and audio information can effectively improve the performances on both tasks.

Figure~\ref{fig:b} presents some failure cases on QVHighlights \cite{lei2021qvhighlights}. Figure~\ref{fig:b}~(a) shows that our model fails to comprehend the time adverbial clause `after a tiring trip'. Instead, it pays more attention to `a young mother and her family' and predicts irrelevant moments. In Figure~\ref{fig:b}~(b), the visual appearances of the retrieved moment and the ground truth are similar. There are few visual clues that can be used to separate `shot' and other actions. Figure~\ref{fig:b}~(c) indicates that our model understands nouns, but can not comprehend abstract words well. We argue that most failure cases are caused by the incomplete understanding of text queries, which may be remitted by using a stronger language model.

\begin{figure*}
\centering
\includegraphics[width=0.975\linewidth]{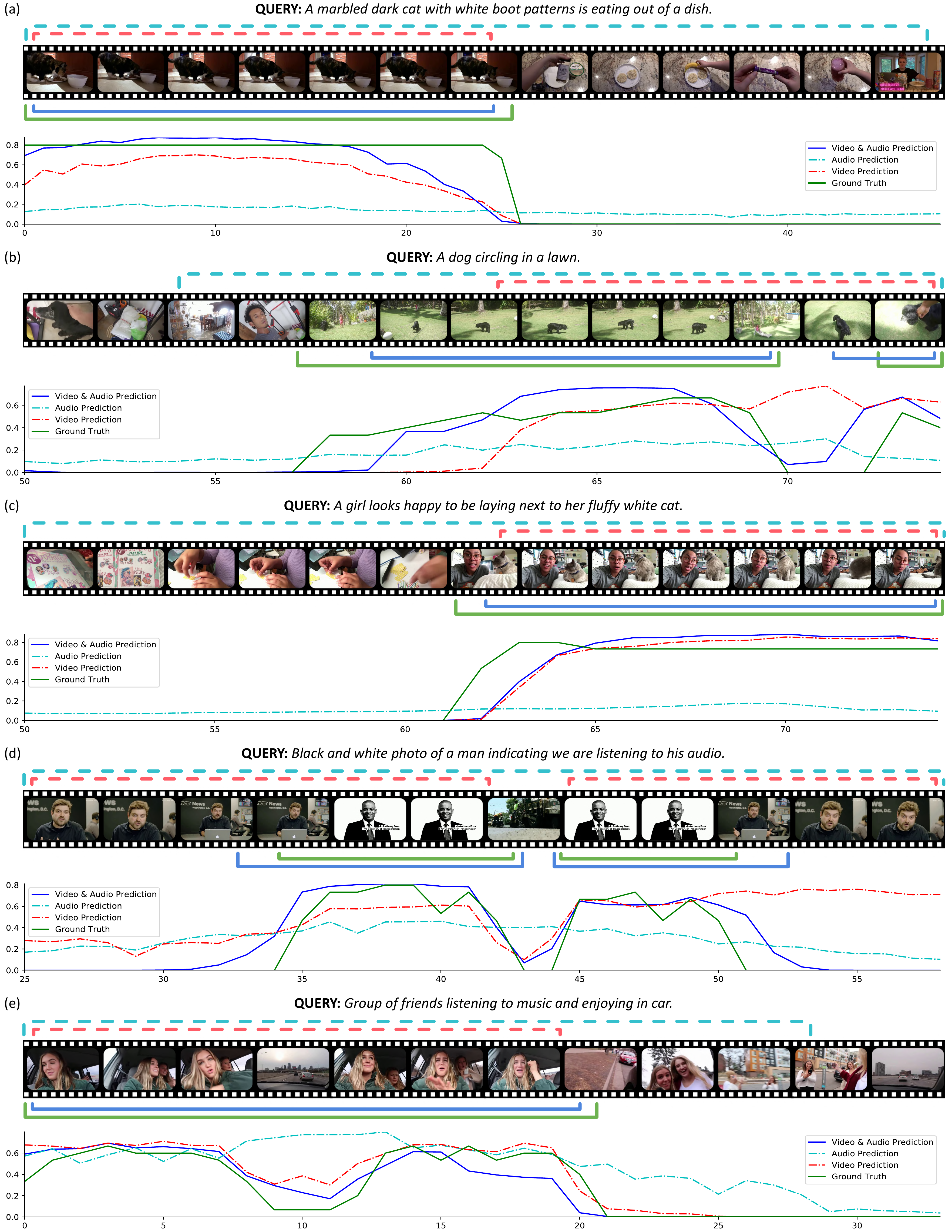}
\caption{Visualization results on QVHighlights \texttt{val} split.}
\label{fig:a}
\end{figure*}

\begin{figure*}
\centering
\includegraphics[width=0.975\linewidth]{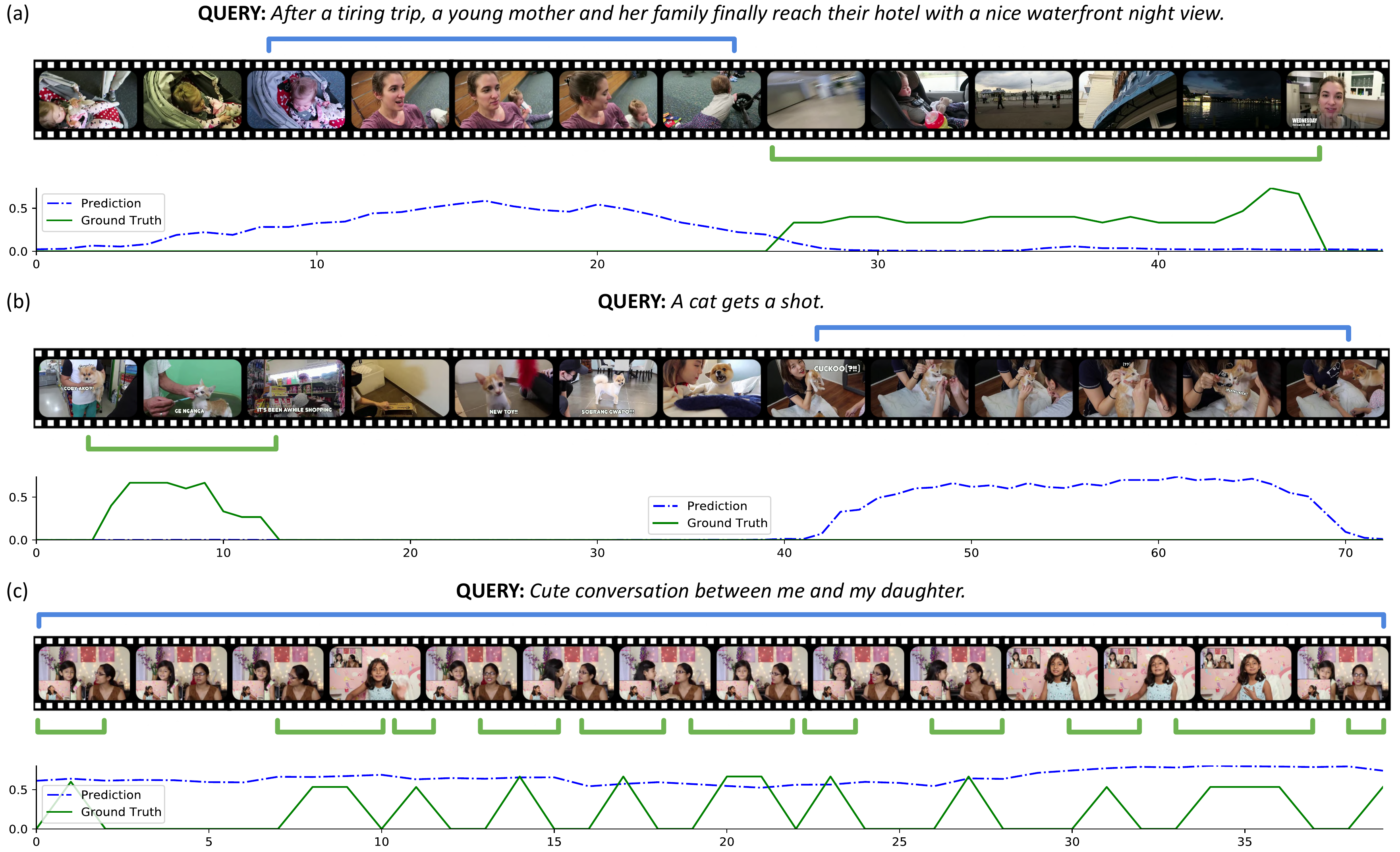}
\caption{Failure cases on QVHighlights \texttt{val} split.}
\label{fig:b}
\end{figure*}}

\end{document}